\documentclass[runningheads]{llncs}

\usepackage{pifont}
\usepackage{amssymb, amsmath, amsfonts}
\usepackage{euscript}
\usepackage{mathtools}
\usepackage{subcaption}
\usepackage{graphicx}
\usepackage{caption}
\usepackage{booktabs}

\usepackage{comment}
\usepackage{graphicx}
\usepackage[table]{xcolor}
\usepackage{hyperref}
\graphicspath{Figures/}
\usepackage{multirow}

\hypersetup{
    colorlinks=true,
    linkcolor=blue,
    filecolor=cyan,      
    urlcolor=magenta,
    }
\begin{document}

\title{ Laplacian-Former: Overcoming the Limitations of Vision Transformers in Local Texture Detection}

\author{Reza Azad\inst{1}\and
Amirhossein Kazerouni\inst{2}\and
Babak Azad\inst{3} \and
Ehsan {Khodapanah Aghdam} \inst{4}\and
Yury Velichko \inst{5}\and
Ulas Bagci \inst{5}\and
Dorit Merhof\inst{6}}


\institute{Faculty of Electrical Engineering and Information Technology, RWTH Aachen University, Germany \and
School of Electrical Engineering, Iran University of Science and Technology, Iran \and
South Dakota State University, Brookings, USA \and
Department of Electrical Engineering, Shahid Beheshti University, Iran \and
Department of Radiology, Northwestern University, Chicago, USA \and
Faculty of Informatics and Data Science, University of Regensburg, Germany
\\\email{azad@pc.rwth-aachen.de}} 

\authorrunning{Azad et al.}
\titlerunning{Laplacian-Former}

\maketitle              

\begin{abstract}
Vision Transformer (ViT) models have demonstrated a breakthrough in a wide range of computer vision tasks. However, compared to the Convolutional Neural Network (CNN) models, it has been observed that the ViT models struggle to capture high-frequency components of images, which can limit their ability to detect local textures and edge information. As abnormalities in human tissue, such as tumors and lesions, may greatly vary in structure, texture, and shape, high-frequency information such as texture is crucial for effective semantic segmentation tasks. To address this limitation in ViT models, we propose a new technique, Laplacian-Former, that enhances the self-attention map by adaptively re-calibrating the frequency information in a Laplacian pyramid. More specifically, our proposed method utilizes a dual attention mechanism via efficient attention and frequency attention while the efficient attention mechanism reduces the complexity of self-attention to linear while producing the same output, selectively intensifying the contribution of shape and texture features. Furthermore, we introduce a novel efficient enhancement multi-scale bridge that effectively transfers spatial information from the encoder to the decoder while preserving the fundamental features. We demonstrate the efficacy of Laplacian-former on multi-organ and skin lesion segmentation tasks with +1.87\% and +0.76\% dice scores compared to SOTA approaches, respectively. Our implementation is publically available at \href{https://github.com/mindflow-institue/Laplacian-Former}{GitHub}.

\end{abstract}

\section{Introduction}
The recent advancements in Transformer-based models have revolutionized the field of natural language processing and have also shown great promise in a wide range of computer vision tasks \cite{chen2021transunet,karimijafarbigloo2023ms}. As a notable example, the Vision Transformer (ViT) model utilizes Multi-head Self-Attention (MSA) blocks to globally model the interactions between semantic tokens created by treating local image patches as individual elements \cite{dosovitskiy2021an}. This approach stands in contrast to CNNs, which hierarchically increase their receptive field from local to global to capture a global semantic representation. Nevertheless, recent studies \cite{wanganti,bai2022improving} have shown that ViT models struggle to capture high-frequency components of images, which can limit their ability to detect local textures and it is vital for many diagnostic and prognostic tasks. This weakness in local representation can be attributed to the way in which ViT models process images. ViT models split an image into a sequence of patches and model their dependencies using a self-attention mechanism, which may not be as effective as the convolution operation used in CNN models in extracting local features within receptive fields. This difference in how ViT and CNN process images may explain the superior performance of CNN models in local feature extraction \cite{geirhosimagenet,azad2021texture}. Innovative approaches have been proposed in recent years to address the insufficient local texture representation within Transformer models. One such approach is the integration of CNN and ViT features through complementary methods, aimed at seamlessly blending the strengths of both in order to compensate for any shortcomings in local representation \cite{chen2021transunet,karimijafarbigloo2023self}.

\textbf{Transformers as a Complement to CNNs:} TransUNet \cite{chen2021transunet} is one of the earliest approaches incorporating the Transformer layers into the CNN bottleneck to model both local and global dependency using the combination of CNN and ViT models. Heidari et al. \cite{heidari2023hiformer} proposed a novel solution called HiFormer, which leverages a Swin Transformer module and a CNN-based encoder to generate two multi-scale feature representations, which are then integrated via a Double-Level Fusion module. UNETR \cite{hatamizadeh2022unetr} used a Transformer to create a powerful encoder with a CNN decoder for 3D medical image segmentation. By bridging the CNN-based encoder and decoder with the Transformer, CoTr \cite{xie2021cotr}, and TransBTS \cite{wang2021transbts}, the segmentation performance in low-resolution stages was improved. Despite these advances, there remain some limitations in these methods such as computationally inefficiency (e.g., TransUNet model), the requirement of a heavy CNN backbone (e.g., HiFormer), and the lack of consideration for multi-scale information. These limitations have resulted in less effective network learning results in the field of medical image segmentation.

\textbf{New Attention Models:} The redesign of the self-attention mechanism within pure Transformer models is another method aiming to augment feature representation to enhance the local feature representation ultimately \cite{karimijafarbigloo2023mmcformer}. In this direction, Swin-Unet \cite{cao2021swin} utilizes a linear computational complexity Swin Transformer \cite{liu2021swin} block in a U-shaped structure as a multi-scale backbone. MISSFormer \cite{huang2021missformer} besides exploring the Efficient Transformer \cite{xie2021segformer} counterpart to diminish the parameter overflow of vision transformers, applies a non-invertible down-sampling operation on input blocks transformer to reduce the parameters. D-Former \cite{wu2022d} is a  pure transformer-based pipeline that comprises a double attention module to capture locally fine-grained attention and interaction with different units in a dilated manner through its mechanism. 

\textbf{Drawbacks of Transformers:} Recent research has revealed that traditional self-attention mechanisms, while effective in addressing local feature discrepancies, have a tendency to overlook important high-frequency information such as texture and edge details \cite{wang2022antioversmooth}. This is especially problematic for tasks like tumor detection, cancer-type identification through radiomics analysis, as well as treatment response assessment, where abnormalities often manifest in texture. Moreover, self-attention mechanisms have a quadratic computational complexity and may produce redundant features\cite{shen2021efficient}. 

\textbf{Our Contributions:} \ding{202} We propose Laplacian-Former, a novel approach that includes new efficient attention (EF-ATT) consisting of two sub-attention mechanisms: \textit{efficient attention} and \textit{frequency attention}. The efficient attention mechanism reduces the complexity of self-attention to linear while producing the same output. The frequency attention mechanism is modeled using a Laplacian pyramid to emphasize each frequency information's contribution selectively. Then, a parametric frequency attention fusion strategy to balance the importance of shape and texture features by recalibrating the frequency features. These two attention mechanisms work in parallel. 
\ding{203} We also introduce a novel efficient enhancement multi-scale bridge that effectively transfers spatial information from the encoder to the decoder while preserving the fundamental features. \ding{204} Our method not only alleviates the problem of the traditional self-attention mechanism mentioned above, but also it surpasses all its counterparts in terms of different evaluation metrics for the tasks of medical image segmentation.


\begin{figure}[!t]
    \centering
    \includegraphics[width=\textwidth]{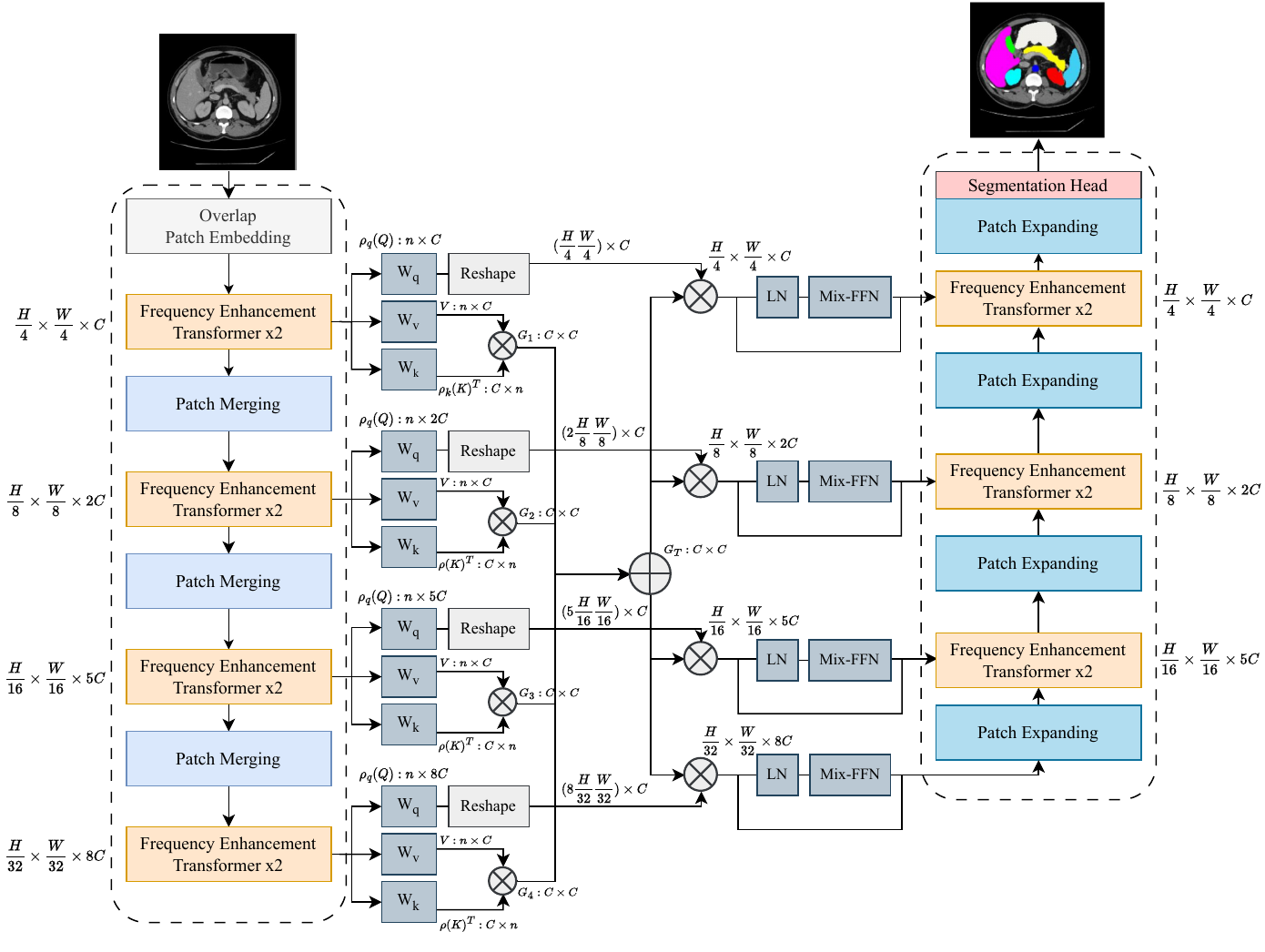}
    \caption{Architecture of our proposed Laplacian-Former.}
    \label{fig:proposed_method}
\end{figure}

\section{Methods}
In our proposed network, illustrated in \autoref{fig:proposed_method}, taking an input image $X \in R^{H\times W \times C}$ with spatial dimensions $H$ and $W$, and $C$ channels, it is first passed through a patch embedding module to obtain overlapping patch tokens of size $4 \times 4$ from the input image. The proposed model comprises four encoder blocks, each containing two efficient enhancement Transformer layers and a patch merging layer that downsamples the features by merging 2×2 patch tokens and increasing the channel dimension. The decoder is composed of three efficient enhancement Transformer blocks and four patch-expanding blocks, followed by a segmentation head to retrieve the final segmentation map. Laplacian-Former then employs a novel efficient enhancement multi-scale bridge to capture local and global correlations of different scale features and effectively transfer the underlying features from the encoder to the decoder.
\begin{figure}[!thb]
    \centering
    \includegraphics[width=\textwidth]{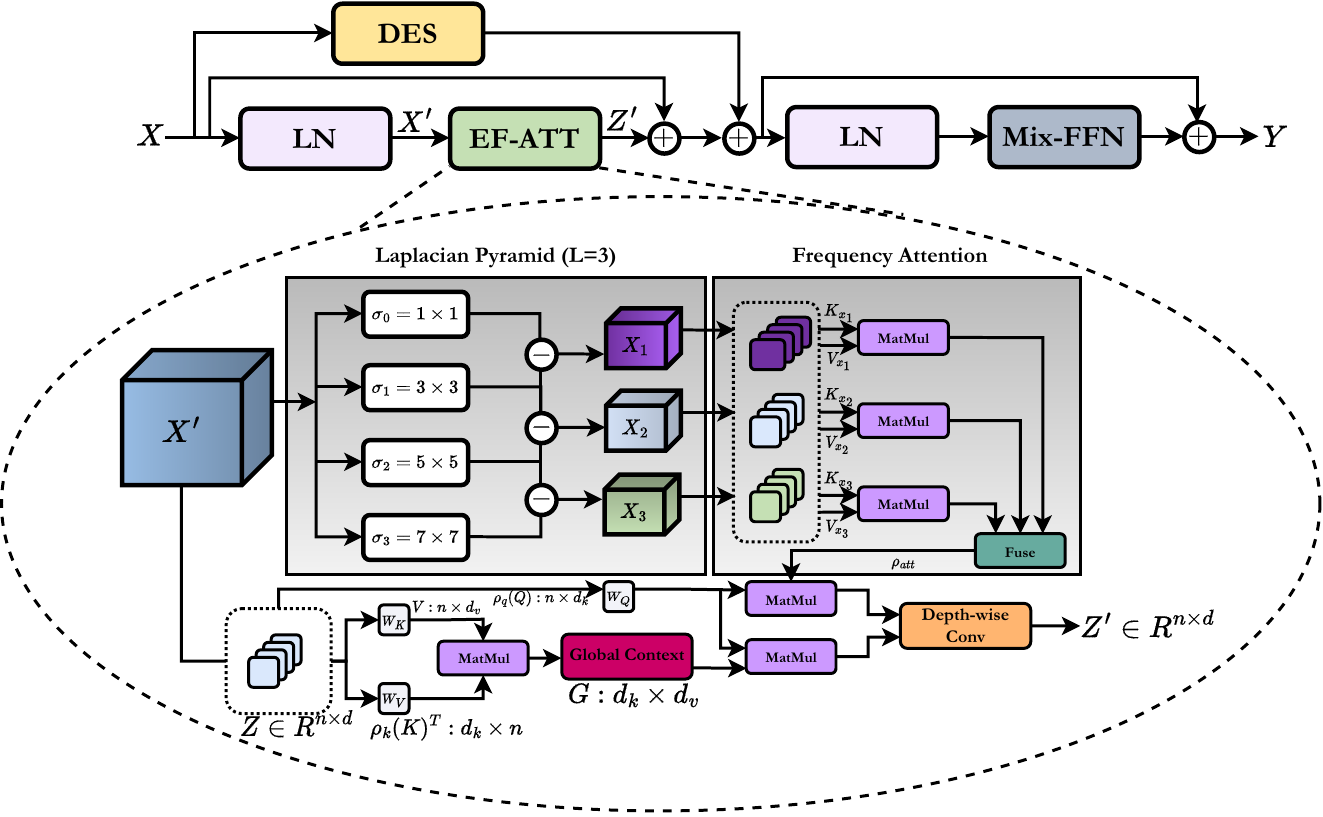}
    \caption{The structure of our frequency enhancement Transformer block.}
    \label{fig:attention}
\end{figure}
\subsection{Efficient Enhancement Transformer Block}
In medical imaging, it is important to distinguish different structures and tissues, especially when tissue boundaries are ill-defined. This is often the case for accurate segmentation of small abnormalities, where high-frequency information plays a critical role in defining boundaries by capturing both textures and edges. Inspired by this, we propose an Efficient Enhancement Transformer Block that incorporates an Efficient Frequency Attention (EF-ATT) mechanism to capture contextual information of an image while recalibrating the representation space within an attention mechanism and recovering high-frequency details. 

Our efficient enhancement Transformer block first takes a LayerNorm (LN) from the input $x$. Then it applies the EF-ATT mechanism to capture contextual information and selectively include various types of frequency information while using the Laplacian pyramid to balance the importance of shape and texture features. Next, $x$ and diversity-enhanced shortcuts are added to the output of the attention mechanism to increase the diversity of features. It is proved in \cite{tang2021augmented} that as Transformers become deeper, their features become less varied, which restrains their representation capacity and prevents them from attaining optimal performance. To address this issue, we have implemented an \textit{augmented shortcut} method from \cite{gu2022multi}, a Diversity-Enhanced Shortcut (DES), employing a Kronecker decomposition-based projection. This approach involves inserting additional paths with trainable parameters alongside the original shortcut $x$, which enhances feature diversity and improves performance while requiring minimal hardware resources. Finally, we apply LayerNorm and MiX-FFN \cite{xie2021segformer} to the resulting feature representation to enhance its power. This final step completes our efficient enhancement Transformer block, as illustrated in \autoref{fig:attention}.
\subsection{Efficient Frequency Attention (EF-ATT)}
The traditional self-attention block computes the attention score $S$ using query ($\mathbf{Q}$) and key ($\mathbf{K}$) values, normalizes the result using Softmax, and then multiplies the normalized attention map with value ($\mathbf{V}$):
\begin{equation}
    S(\mathbf{Q}, \mathbf{K}, \mathbf{V})=Softmax\left(\frac{\mathbf{Q K}^{\mathbf{T}}}{\sqrt{d_k}}\right) \mathbf{V},
\end{equation}
where $d_k$ is the embedding dimension. One of the main limitations of the dot-product mechanism is that it generates redundant information, resulting in unnecessary computational complexity. Shen et al. \cite{shen2021efficient} proposed to represent the context more effectively by reducing the computational burden from $\mathcal{O}(n^2)$ to linear form $\mathcal{O}(d^2n)$:
\begin{equation}
    E(\mathbf{Q}, \mathbf{K}, \mathbf{V})=\rho_{\mathbf{q}}(\mathbf{Q})\left(\rho_{\mathbf{k}}(\mathbf{K})^{\mathbf{T}} \mathbf{V}\right).
\end{equation}

Their approach involves applying the Softmax function ($\rho$) to the key and query vectors to obtain normalized scores and formulating the global context by multiplying the key and value matrix. They demonstrate that efficient attention $E$ can provide an equivalent representation of self-attention while being computationally efficient. By adopting this approach, we can alleviate the issues of feature redundancy and computational complexity associated with self-attention.

Wang et al. \cite{wang2022antioversmooth} explored another major limitation of the self-attention mechanism, where they demonstrated through theoretical analysis that self-attention operates as a low-pass filter that erases high-frequency information, leading to a loss of feature expressiveness in the model's deep layers. Authors found that the Softmax operation causes self-attention to keep low-frequency information and loses its fine details. Motivated by this, we propose a new frequency recalibration technique to address the limitations of self-attention, which only focuses on low-frequency information (which contains shape information) while ignoring the higher frequencies that carry texture and edge information. First, we construct a Laplacian pyramid to determine the different frequency levels of the feature maps. The process begins by extracting $(L + 1)$ Gaussian representations from the encoded feature using different variance values of the Gaussian function:
\begin{equation}
    \mathbf{G}_l(\mathbf{X})=\mathbf{X} * \frac{1}{\sigma_l \sqrt{2 \pi}} e^{-\frac{i^2+j^2}{2 \sigma_l^2}},
\end{equation}
where $\mathbf{X}$ refers to the input feature map, $(i, j)$ corresponds to the spatial location within the encoded feature map, the variable $\sigma_l$ denotes the variance of the Gaussian function for the $l$-th scale, and the symbol $*$ represents the convolution operator. The pyramid is then built by subtracting the $l$-th Gaussian function ($\mathbf{G}_l$) output from the $(l+1)$-th output ($\mathbf{G}_l-\mathbf{G}_{l+1}$) to encode frequency information at different scales.
The Laplacian pyramid is composed of multiple levels, each level containing distinct types of information. To ensure a balanced distribution of low and high-frequency information in the model, it is necessary to efficiently aggregate the features from all levels of the frequency domain. Hence, we present frequency attention that involves multiplying the key and value of each level $(\mathbf{X}_l)$ to calculate the attention score and then fuses the resulting attention scores of all levels using a fusion module, which performs summation. The resulting attention score is multiplied by Query $(\mathbf{Q})$ to obtain the final frequency attention result, which subsequently concatenates with the efficient attention result and applies the depth-wise convolution with the kernel size of $2\times1\times1$ in order to aggregate both information and recalibrate the feature map, thus allowing for the retrieval of high-frequency information.
\subsection{Efficient Enhancement Multi-scale Bridge}
It is widely known that effectively integrating multi-scale information can lead to improved performance \cite{huang2021missformer}. Thus, we introduce the Efficient Enhancement Multi-scale Bridge as an alternative to simply concatenating the features from the encoder and decoder layers. The proposed bridge, depicted in \autoref{fig:proposed_method}, delivers spatial information to each decoder layer, enabling the recovery of intricate details while generating output segmentation masks. In this approach, we aim to calculate the efficient attention mechanism for each level and fuse the multi-scale information in their context; thus, it is important that all levels' embedding dimension is of the same size. Therefore, in order to calculate the global context $(\mathbf{G}_i)$, we parametrize the query and value of each level using a convolution $1\times1$ where it gets the size of $mC$ and outputs $C$, where $m$ equals 1, 2, 5, and 8 for the first to fourth levels, respectively. We multiply the new key and value to each other to attain the global context. We then use a summation module to aggregate the global context of all levels and reshape the query for matrix multiplication with the augmented global context. Taking the second level with the dimension of $\frac{H}{8}\times \frac{W}{8}\times 2C$, the key and value are mapped to $(\frac{H}{8}\frac{W}{8}) \times C$, and the query to $(2\frac{H}{8}\frac{W}{8}) \times C$. The augmented global context with the shape of $C\times C$ is then multiplied by the query, resulting in an enriched feature map with the shape of $(2\frac{H}{8}\frac{W}{8}) \times C$.  We reshape the obtained feature map into $\frac{H}{8}\times \frac{W}{8} \times 2C$ and feed it through an LN and MiX-FFN module with a skip connection to empower the feature representations. The resulting output is combined with the expanded feature map, and then projected using a linear layer onto the same size as the encoder block corresponding to that level.


\section{Results}
Our proposed technique was developed using the PyTorch library and executed on a single RTX 3090 GPU. A batch size of 24 and a stochastic gradient descent algorithm with a base learning rate of 0.05, a momentum of 0.9, and a weight decay of 0.0001 was utilized during the training process, which was carried out for 400 epochs. For the loss function, we used both cross-entropy and Dice losses ($Loss = \gamma \cdot L_{dice} + (1-\gamma) \cdot L_{ce}$), $\gamma$ set to $0.6$ empirically.

\noindent\textbf{Datasets:}
We tested our model using the \textit{Synapse} dataset \cite{landman2015miccai}, which comprises 30 cases of contrast-enhanced abdominal clinical CT scans (a total of 3,779 axial slices). Each CT scan consists of $85 \sim 198$ slices of the in-plane size of $512 \times 512$ and has annotations for eight different organs.
We followed the same preferences for data preparation analogous to \cite{chen2021transunet}. We also followed \cite{azad2022transdeeplab,eskandari2023inter} experiments to evaluate our method on the ISIC 2018 skin lesion dataset \cite{codella2019skin} with 2,694 images. 
%
\begin{table*}[!t]
    \centering
    \caption{Comparison results of the proposed method on the \textit{Synapse} dataset. \textcolor{blue}{Blue} indicates the best result, and \textcolor{red}{red} indicates the second-best.}
    \resizebox{\textwidth}{!}{
        \begin{tabular}{l|c|cc|cccccccc}
            \toprule
            \textbf{Methods}        &   \textbf{\# Params (M)}             & \textbf{DSC~$\uparrow$} & \textbf{HD~$\downarrow$} & \textbf{Aorta}          & \textbf{Gallbladder}    & \textbf{Kidney(L)}      & \textbf{Kidney(R)}      & \textbf{Liver}          & \textbf{Pancreas}       & \textbf{Spleen}         & \textbf{Stomach}        \\
            \midrule
            R50 U-Net \cite{chen2021transunet} &  30.42  & 74.68  & 36.87 & 87.74 & 63.66 & 80.60 & 78.19 & 93.74 & 56.90 & 85.87 & 74.16
            \\
            U-Net \cite{ronneberger2015unet} & 14.8  & 76.85 & 39.70 & \textcolor{red}{89.07}  & \textcolor{red}{69.72}  & 77.77 & 68.60    & 93.43 & 53.98 & 86.67 & 75.58
            \\
            Att-UNet \cite{schlemper2019attention} &  34.9 & 77.77 & 36.02 & \textcolor{blue}{89.55} & 68.88 & 77.98 & 71.11               & 93.57  & 58.04 & 87.30 & 75.75
            \\
            TransUNet \cite{chen2021transunet}   &  105.28  & 77.48                   & 31.69                    & 87.23                   & 63.13                   & 81.87                   & 77.02                   & 94.08                   & 55.86                   & 85.08                   & 75.62
            \\
            Swin-Unet \cite{cao2021swin}    &   27.17     & 79.13                   & 21.55                    & 85.47                   & 66.53                   & 83.28                   & 79.61                   & 94.29                   & 56.58                   & 90.66                   & 76.60
            \\
            LeVit-Unet \cite{xu2021levit}    &  52.17       & 78.53                   & \textcolor{red}{16.84}                    & 78.53                   & 62.23                   & 84.61                   & 80.25                   & 93.11                   & 59.07                   & 88.86                   & 72.76
            \\
            TransDeepLab \cite{azad2022transdeeplab} &  21.14  & 80.16 & 21.25 & 86.04 & 69.16 &84.08 & 79.88 & 93.53 & 61.19 & 89.00 & 78.40 \\
            HiFormer \cite{heidari2023hiformer}   &   25.51  & 80.39                   & \textcolor{blue}{14.70}  & 86.21                   & 65.69                   & \textcolor{blue}{85.23}  & 79.77                   & 94.61                   & 59.52                   & 90.99                   & \textcolor{red}{81.08}
            \\
            \hline
            EffFormer           &    22.31    & 80.79                   & 17.00   & 85.81                   & 66.89                   & 84.10                   & \textcolor{blue}{81.81} & \textcolor{red}{94.80}  & 62.25                   & 91.05                   & 79.58
            \\
            LaplacianFormer (without bridge) &  23.87  &  \textcolor{red}{81.59} & 17.31  & 87.41  & 69.57  & \textcolor{red}{85.22} & 80.46 & 94.68 & \textcolor{red}{63.71}  & \textcolor{red}{91.47} & 78.23
            \\
            \rowcolor{cyan!10}
            \textbf{LaplacianFormer} &  27.54  &\textcolor{blue}{81.90} & 18.66                    & 86.55 & \textcolor{blue}{71.19} & 84.23 & \textcolor{red}{80.52}  & \textcolor{blue}{94.90} & \textcolor{blue}{64.75}  & \textcolor{blue}{91.91} & \textcolor{blue}{81.14}
            \\
            \bottomrule
        \end{tabular}
    }\label{tab:synapse_performance_comparison}
\end{table*}

\begin{figure}[!t]
    \centering
    \includegraphics[width=\textwidth]{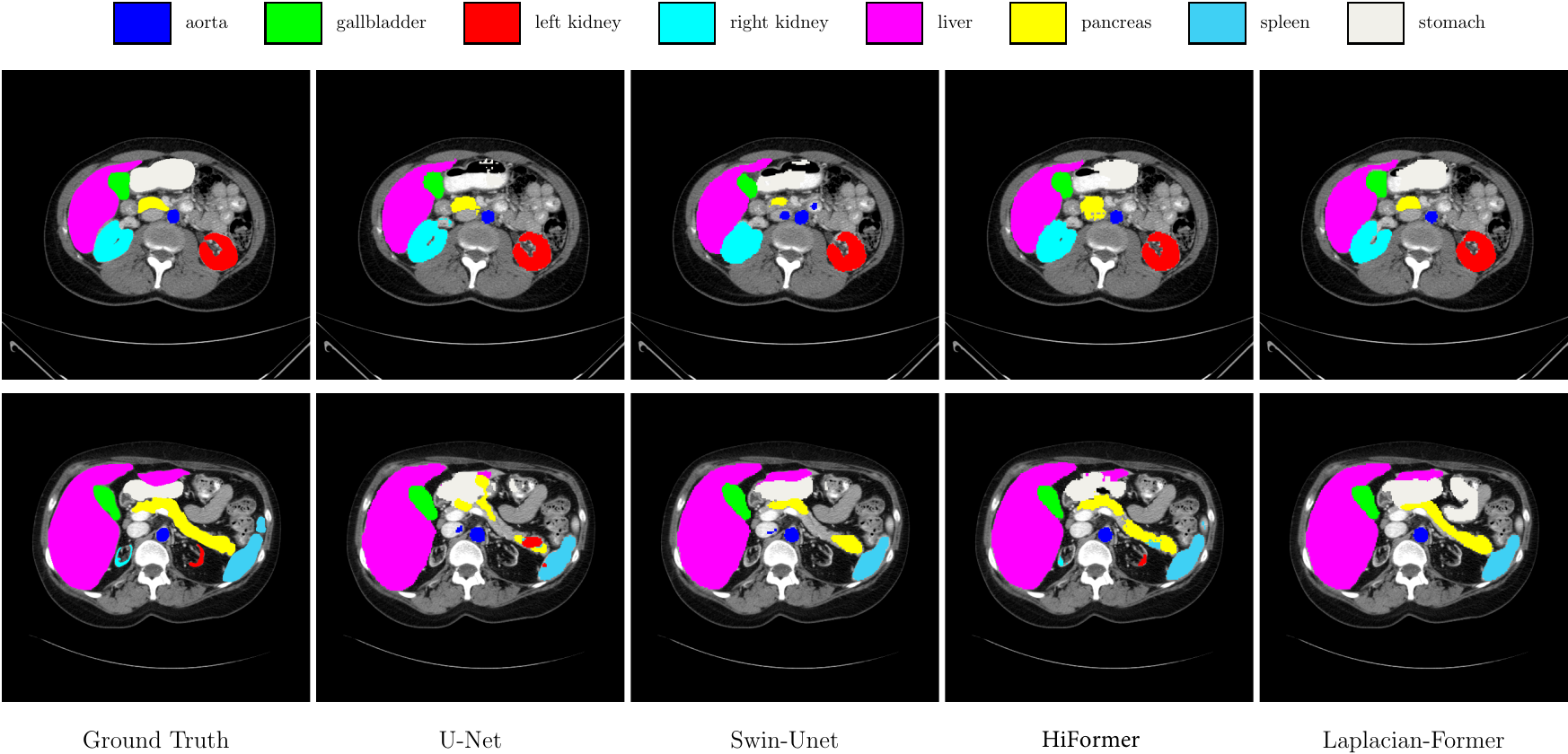}
    \caption{Segmentation results of the proposed method on the \textit{Synapse} dataset. Our Laplacian-Former shows finer boundaries (high-frequency details) for the region of the stomach and less false positive prediction for the pancreas.}
	\label{fig:synapseviz}
\end{figure}

\noindent\textbf{Synapse Multi-Organ Segmentation:}
\autoref{tab:synapse_performance_comparison} presents a comparison of our proposal with previous SOTA methods using the DSC and HD metrics across eight abdominal organs. Laplacian-Former clearly outperforms SOTA CNN-based methods. We extensively evaluated EfficientFormer (EffFormer) plus another drift of Laplacian-Former without utilizing the bridge connections to endorse the superiority of Laplacian-Former. Laplacian-Former exhibits superior learning ability on the Dice score metric compared to other transformer-based models, achieving an increase of +$1.59\%$ and +$2.77\%$ in Dice scores compared to HiFormer and Swin-Unet, respectively. \autoref{fig:synapseviz} illustrates a qualitative result of our method for different organ segmentation, specifically we can observe that the LalacianFormer produces a precise boundary segmentation on Gallbladder, Liver, and Stomach organs. It is noteworthy to mention that our pipeline, as a pure transformer-based architecture trained from scratch without pretraining weights, outperforms all previously presented network architectures.

\noindent\textbf{Skin Lesion Segmentation:}
\autoref{tab:skin_comparison} shows the comparison results of our proposed method, Laplacian-Former, against leading methods on the skin lesion segmentation benchmark. Our approach outperforms other competitors across most evaluation metrics, indicating its excellent generalization ability across different datasets. In particular, our approach performs better than hybrid methods such as TMU-Net \cite{reza2022contextual} and pure transformer-based methods such as Swin-Unet \cite{cao2021swin}. Our method achieves superior performance by utilizing the frequency attention in a pyramid scale to model local textures. Specifically, our frequency attention emphasizes the fine details and texture characteristics that are indicative of skin lesion structures and amplifies regions with significant intensity variations, thus accentuating the texture patterns present in the image and resulting in better performance. In addition, we provided the spectral response of LaplacianFormer vs. Standard Transformer in identical layers in \autoref{fig:spectralResponse}. It is evident Standard design frequency response in deep layers of structure attenuates more than the LaplacianFormer, which is a visual endorsement of the capability of LaplacianFormer for its ability to preserve high-frequency details. The supplementary provides more visualization results. 
\begin{table}[!thb]
    \caption{(a) Performance comparison of Laplacian-Former against the SOTA approaches on \textit{ISIC 2018} skin lesion datset. \textcolor{blue}{Blue} and \textcolor{red}{red} indicates the best and the second-best results. (b) Frequency response analysis on the LaplacianFormer (up) vs. Standard Transformer (down).}
    \begin{subtable}[!h]{0.47\textwidth}
    \centering
    \caption{\textit{ISIC 2018} dataset}
    \label{tab:skin_comparison}
    \resizebox{\textwidth}{!}{
    \begin{tabular}{l||cccc} 
    \toprule
    \multirow{2}{*}{\textbf{Methods}}  &  \multicolumn{4}{c}{\textbf{ISIC 2018}} \\ 
    \cline{2-5}
     & \textbf{DSC} & \textbf{SE} & \textbf{SP} & \textbf{ACC} \\ 
    \midrule
    U-Net \cite{ronneberger2015u} & 0.8545 & 0.8800 & 0.9697 & 0.9404  
    \\
    Att-UNet \cite{schlemper2019attention} & 0.8566 & 0.8674 & \textcolor{blue}{0.9863} & 0.9376 
    \\
    TransUNet \cite{chen2021transunet} & 0.8499 & 0.8578 & 0.9653 & 0.9452  
    \\
    FAT-Net \cite{wu2022fat}  & 0.8903 & \textcolor{red}{0.9100} & 0.9699 & 0.9578 
    \\
    TMU-Net \cite{reza2022contextual}  & 0.9059 & 0.9038 & 0.9746 & 0.9603  
    \\
    Swin-Unet \cite{cao2021swin}  & 0.8946 & 0.9056 & \textcolor{red}{0.9798} & 0.9605
    \\
    \midrule
    EffFormer & 0.8909 & 0.9034 & 0.9701 & 0.9579 
    \\
    Laplacian-Former (without bridge) & \textcolor{red}{0.9100} & 0.9289 & 0.9655 & \textcolor{red}{0.9611} 
    \\
    \rowcolor[HTML]{C8FFFD}
    \textbf{Laplacian-Former} & \textcolor{blue}{0.9128} & \textcolor{blue}{0.9290} & 0.9715 & \textcolor{blue}{0.9626} 
    \\
    \bottomrule
    \end{tabular}
    }
    \end{subtable}
    \begin{subtable}[!h]{0.47\textwidth}
    \centering
    \caption{Spectral Response} \label{fig:spectralResponse}
    \begin{tabular}{c}
         \includegraphics[width=\textwidth]{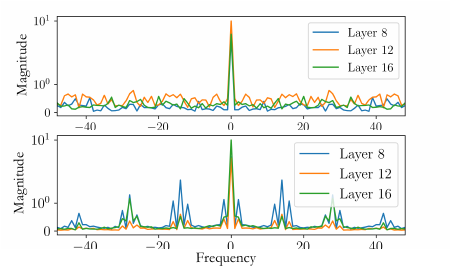}
    \end{tabular}
    \end{subtable}
\end{table}
\section{Conclusion}
In this paper, we introduce Laplacian-Former, a novel standalone transformer-based U-shaped architecture for medical image analysis. Specifically, we address the transformer's inability to capture local context as high-frequency details, e.g., edges and boundaries, by developing a new design within a scaled dot attention block. Our pipeline benefits the multi-resolution Laplacian module to compensate for the lack of frequency attention in transformers. Moreover, while our design takes advantage of the efficiency of transformer architectures, it keeps the parameter numbers low.


\bibliographystyle{splncs04}
\bibliography{ref.bib}

\clearpage
\newpage
\appendix
\newpage

\section*{Supplementary Material}
\begin{figure}[h]
\centering
\resizebox{\textwidth}{!}{
    \begin{tabular}{@{} *{6}c @{}}
    \includegraphics[width=0.25\textwidth]{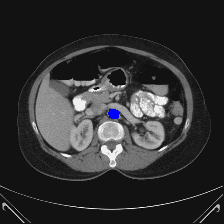} &
    \includegraphics[width=0.25\textwidth]{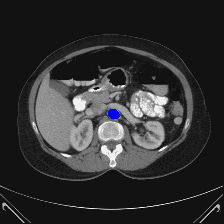} &
    \includegraphics[width=0.25\textwidth]{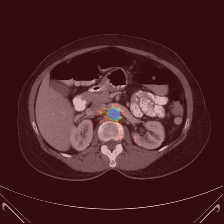} &
    \includegraphics[width=0.25\textwidth]{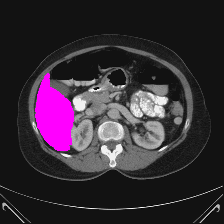} &
    \includegraphics[width=0.25\textwidth]{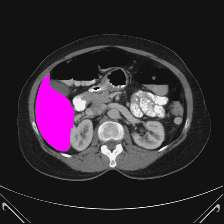} &
    \includegraphics[width=0.25\textwidth]{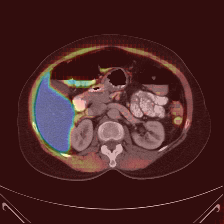} \\
    \includegraphics[width=0.25\textwidth]{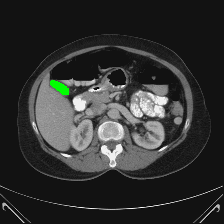} &
    \includegraphics[width=0.25\textwidth]{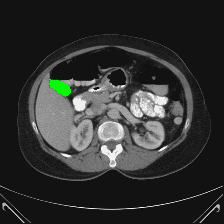} &
    \includegraphics[width=0.25\textwidth]{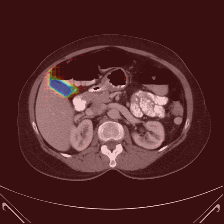} &
    \includegraphics[width=0.25\textwidth]{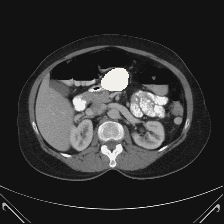} &
    \includegraphics[width=0.25\textwidth]{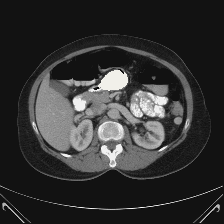} &
    \includegraphics[width=0.25\textwidth]{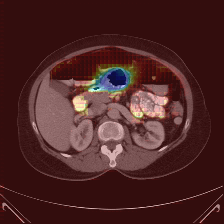} \\
    {\small (a) Ground Truth} & {\small(b) Prediction} & {\small(c) Heatmap} & {\small (d) Ground Truth} & {\small(e) Prediction} & {\small(f) Heatmap}
    \end{tabular}
}
\caption{The LaplacianFormer model's ability to detect organs in the \textit{Synapse} dataset was evaluated through the visualization of its attention map using Grad-CAM. The findings suggest that the LaplacianFormer model is effective in detecting small organs such as the gallbladder and aorta from the right side's top to bottom in order, indicating its effectiveness in learning local features. Additionally, the model effectively detects large organs such as the liver and stomach from the left side's top to bottom, demonstrating its capability to capture long-range dependencies.}
\label{fig:heat_map}
\end{figure}

\begin{figure}
    \centering
    \includegraphics[width=\textwidth]{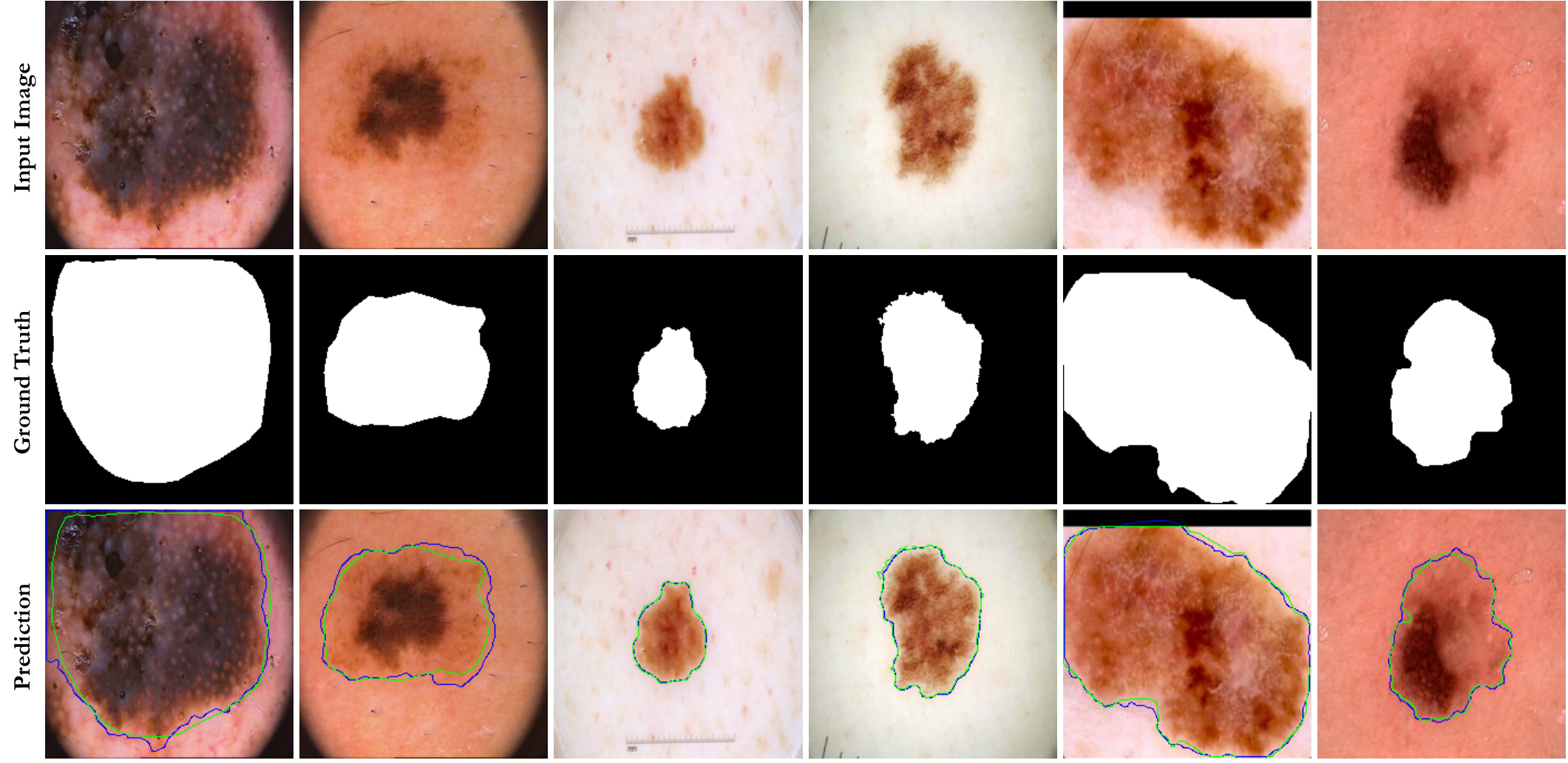}
    \caption{Visual comparisons on the \textit{ISIC2018} skin lesion dataset. Ground truth boundaries are shown in \textcolor{green}{green}, and predicted boundaries are shown in \textcolor{blue}{blue}.}
    \label{fig:my_label}
\end{figure}
\vspace{-1em}
\begin{figure}
    \centering
    \includegraphics[width=1\textwidth]{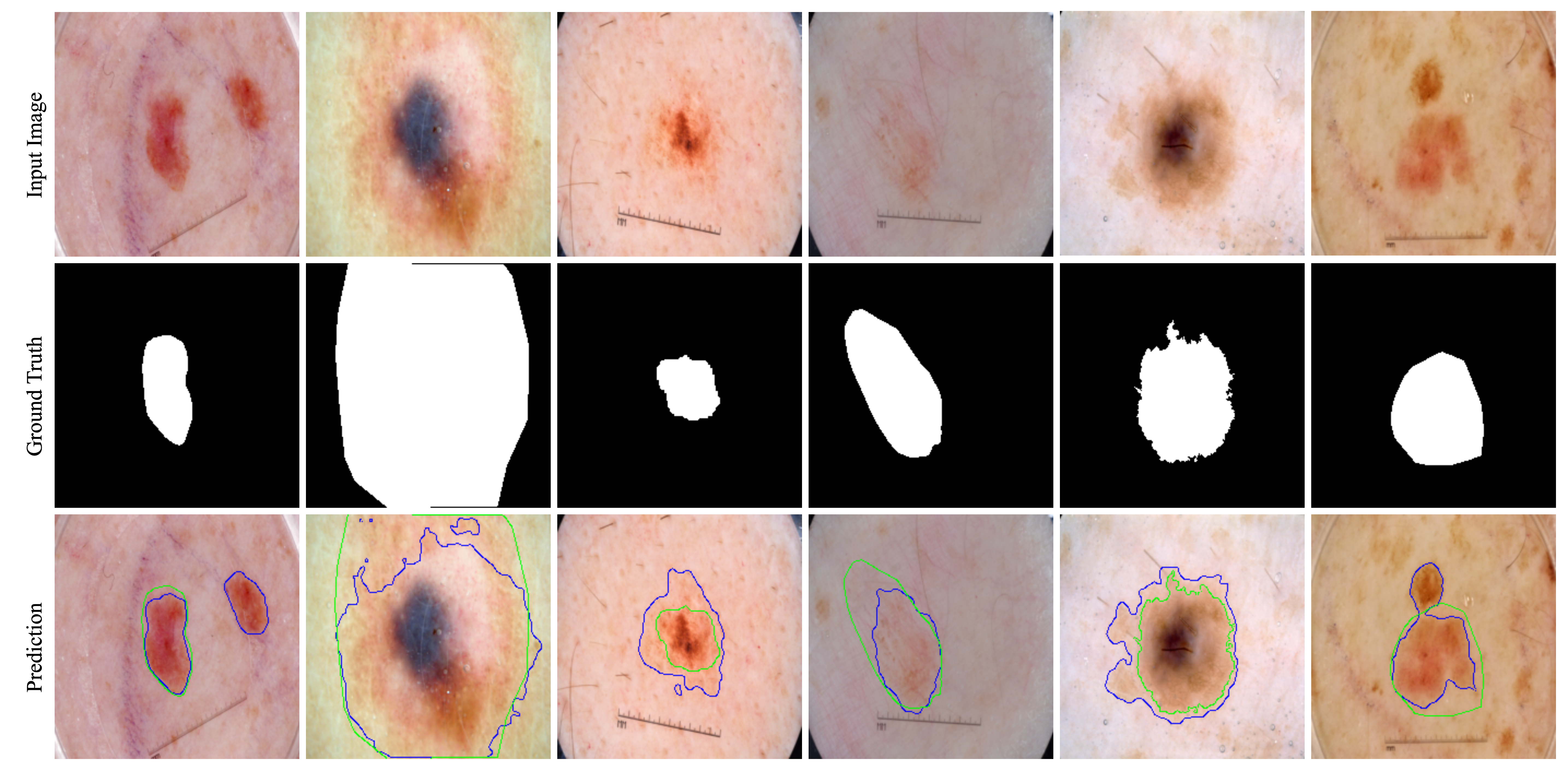}
    \caption{Pitfalls on skin lesion segmentation task with Laplacian-Former. From the first column, the model captured the possible lesions from the image as opposed to the lack o annotation. Column two and three shows that the model failed to capture the boundaries as the ground truth however it performed well in capturing the local texture. In the fourth to the last column, the model performed accurately, while the annotation is coarse rather than fine.}
    \label{fig:skin_pitfalls}
\end{figure}


\end{document}